\documentclass[sigconf]{acmart}

\usepackage{multirow}
\graphicspath{{./figs/}}
\usepackage[caption=false]{subfig}

\makeatletter
\newenvironment{subfigures}
 {\begin{minipage}{\columnwidth}\def\@captype{figure}\centering}
 {\end{minipage}}
\makeatother

\usepackage{color,soul}
\usepackage{colortbl}

\setcopyright{rightsretained}

\begin{document}
\title{Novelty Producing Synaptic Plasticity}


\author{Anil Yaman}
\affiliation{%
  \institution{Eindhoven University of Technology}
  \city{Eindhoven} 
  \state{the Netherlands} 
}
\email{a.yaman@tue.nl}

\author{Giovanni Iacca}
\affiliation{%
  \institution{University of Trento}
  \city{Trento} 
  \state{Italy} 
}
\email{giovanni.iacca@unitn.it}

\author{Decebal Constantin Mocanu}
\affiliation{%
  \institution{Eindhoven University of Technology}
  \city{Eindhoven} 
  \state{the Netherlands} 
}
\email{d.c.mocanu@tue.nl}

\author{George Fletcher}
\affiliation{%
  \institution{Eindhoven University of Technology}
  \city{Eindhoven} 
  \state{the Netherlands} 
}
\email{g.h.l.fletcher@tue.nl}

\author{Mykola Pechenizkiy}
\affiliation{%
  \institution{Eindhoven University of Technology}
  \city{Eindhoven} 
  \state{the Netherlands} 
}
\email{m.pechenizkiy@tue.nl}

\renewcommand{\shortauthors}{A. Yaman et al.}

\begin{abstract}

A learning process with the plasticity property often requires reinforcement signals to guide the process. However, in some tasks (e.g. maze-navigation), it is very difficult (or impossible) to measure the performance of an agent (i.e. a fitness value) to provide reinforcements since the position of the goal is not known. This requires finding the correct behavior among a vast number of possible behaviors without having the knowledge of the reinforcement signals. In these cases, an exhaustive search may be needed. However, this might not be feasible especially when optimizing artificial neural networks in continuous domains. In this work, we introduce novelty producing synaptic plasticity (NPSP), where we evolve synaptic plasticity rules to produce as many novel behaviors as possible to find the behavior that can solve the problem. We evaluate the NPSP on maze-navigation on deceptive maze environments that require complex actions and the achievement of subgoals to complete. Our results show that the search heuristic used with the proposed NPSP is indeed capable of producing much more novel behaviors in comparison with a random search taken as baseline.


\end{abstract}

%
%

\ccsdesc[500]{Theory of computation~Evolutionary algorithms}
\ccsdesc[500]{Theory of computation~Bio-inspired optimization}

\keywords{Unsupervised learning, novelty search, task-agnostic learning, synaptic plasticity}

\maketitle

\section{Introduction}\label{sec:introduction}

During a learning process, the fitness value of each behavior can be measured and used as reinforcement signal to guide the learning process. For instance, in a maze-navigation task, a fitness measure such as the distance of an agent to the goal position can be used as reinforcement to optimize its behavior. However, in realistic scenarios, this fitness measure might not be available since the goal position is not known.

\begin{figure}[H]
\begin{center}
\includegraphics[width=1\columnwidth]{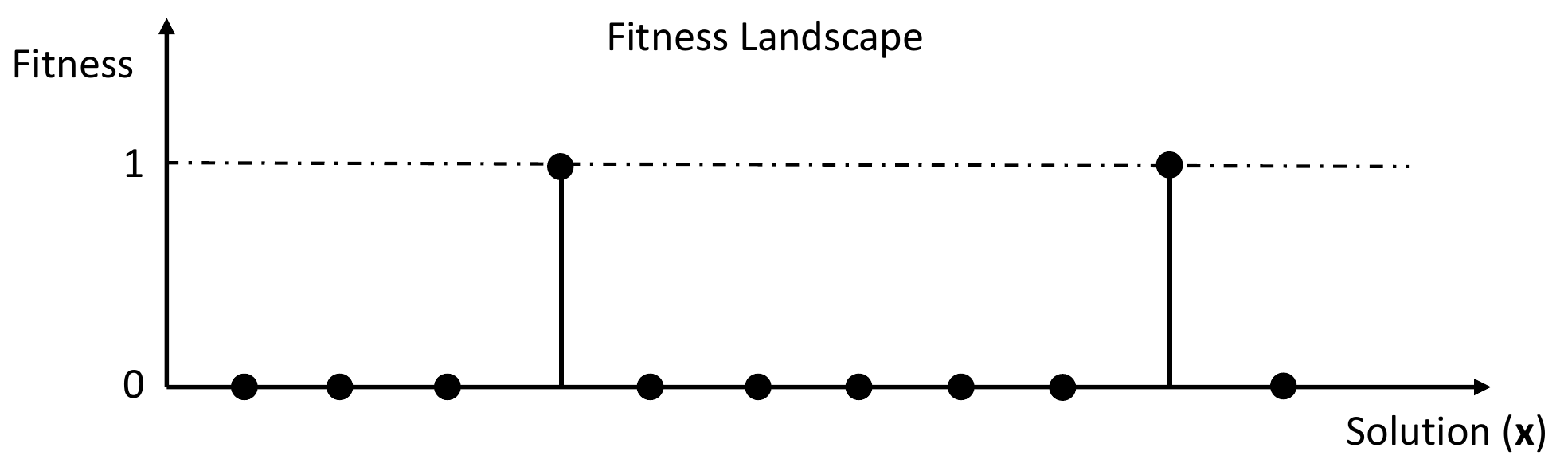}
\caption{An illustration of a hypothetical fitness landscape where there is only two possible discrete fitness values (i.e., 1: global optimum, 0: global minimum). The $x$- and $y$-axes show the candidate solutions and their fitness values respectively. Typically, only a small set of candidate solutions associated with the high fitness value that indicates the solution to the problem. For instance, in the case of maze-navigation, only the behaviors that achieve to the goal position have a fitness value of 1, the rest of the behaviors that fail to achieve to the goal position have a fitness value of 0. Only a small fraction of all possible behaviors can achieve to the goal position.}
\label{fig:fitnessLandscape}
\end{center}
\end{figure}

We consider a learning process where it is very difficult (or impossible) to measure the fitness of a behavior of an agent to provide reinforcement signals. We refer to this problem as the \textit{needle in a haystack} problem~\cite{hinton1996learning} where the \textit{needle} refers to a solution (i.e. a behavior that can solve the task) and the \textit{haystack} refers to the search space (i.e. all possible behaviors).

A hypothetical case of an illustration of the fitness landscape of the needle in a haystack problem is given in Figure~\ref{fig:fitnessLandscape}. The $x$- and $y$-axes show the solutions (behaviors) and their fitness values respectively. The problem assumes that there is no available metric to measure quantitatively the fitness of a behavior: the task is either solved or not. Therefore, fitness values 1 and 0 indicate the successful and failed behaviors. There may be more than one behavior that can provide a solution to the problem; on the other hand, we assume that the majority of the behaviors fail to solve the task.

Novelty search and MAP-Elites algorithms have been successfully used in tasks where the use of fitness values is often detrimental for finding good solutions via traditional (fitness-driven) evolutionary search~\cite{lehman2008exploiting,mouret2015illuminating}. These algorithms may be beneficial for solving the needle in a haystack problems. However, they require external memory to store encountered solutions and, in the case of MAP-Elites, fitness values to map the solutions to a predefined feature space.

In this work, we propose novelty producing synaptic plasticity (NPSP) for the needle in a haystack problem, where we use synaptic plasticity to produce novel behaviors. The synaptic plasticity performs changes in connection weights of artificial neural networks (ANNs) based on the local activation of neurons. We use genetic algorithms to optimize the NPSP rules to produce as many novel behaviors as possible to find the behavior that can solve the task. In contrast to novelty search, the NPSP performs changes in a single ANN (controlling a single agent) without keeping track of produced behaviors.

We evaluate the performance of the NPSP on maze-navigation task using deceptive maze environments which require complex actions and the achievements of subgoals to complete. During the evaluation phase, we assume that the knowledge of the fitness value, in terms of the distance of the agent to the goal position, is not available. Our results show that the proposed NPSP produces a large number of behaviors relative to a random search that may eventually help finding the solution to the problems when the fitness function is not known or difficult to evaluate.

The rest of the paper is organized as follows: in Section~\ref{sec:EvolvingNPSP}, we provide background knowledge on evolution of synaptic plasticity, then we introduce our method to produce novelty producing synaptic plasticity in ANNs. In Section~\ref{sec:ExperimentalSetup}, we provide the details about our experimental setup where we discuss the test environments, agent architecture, genetic algorithm and benchmark algorithms. In Section~\ref{sec:ExperimentalResults} we provide our results, and finally, in Section~\ref{sec:Conclusions}, we discuss our conclusions and future work.


\section{Evolving Plasticity for Producing Novelty} \label{sec:EvolvingNPSP}

The \textit{synaptic plasticity} refers to the property of biological neural networks (BNNs) that allows them to change their configuration during their lifetime. These changes are known to occur in synapses (i.e. connections between neurons) based on the local information~\cite{holtmaat2009experience}. Hebbian learning was proposed to model the synaptic plasticity in ANNs~\cite{hebb1949,kuriscak2015}. In this form of learning, \textit{synaptic plasticity rules} are used to adjust the weight of a connection between two neurons based on the correlations between the neurons prior (pre-synaptic neuron) and posterior (post-synaptic neuron) with respect to the connection. Moreover, reinforcement signals are used to guide the learning process by performing these adjustments in order to match the neuron outputs with the desired behaviors.

The basic form of Hebbian learning can suffer from instability because when an increase in the connection weight between two neurons leads to an increase in their correlations this in turn causes further increase in their connection weights. To reduce this effect, several variants of Hebbian learning rules have been proposed in the literature~\cite{vasilkoski2011review}. Nevertheless, further optimization may be needed to find learning rules that can produce stable and coherent learning for certain learning scenarios.

Inspired by the evolution of learning in BNNs, evolutionary computing has been used to optimize the plasticity rules to produce plasticity property in ANNs~\cite{soltoggio2017born}. A number of previous works optimized the type of Hebbian learning rule and its parameters~\cite{floreano2000evolutionary,niv2002evolution}; some other works used more complex models (i.e. additional ANNs) to perform synaptic changes~\cite{risi2010indirectly,orchard2016evolution}. 

Here, we optimize the synaptic plasticity rules to encourage the novel behaviors. This may especially be beneficial in cases where there is no information (i.e. fitness values, reinforcement signals) about the problem to guide the learning process. 

In an ANN, the activation of a post-synaptic neuron $a_i$ is computed by:
\vspace{-0.1cm}
\begin{eqnarray}
a_i = \psi 
\left( 
\sum_{j=0}  w_{ij} \cdot a_j 
\right) 
\label{eq:activationPost}
\end{eqnarray}
where $a_j$ is the pre-synaptic neuron activation, $w_{ij}$ is the connection weight between pre- and post-synaptic neurons, and $\psi$ is the activation function. We use a step function which assigns 0 to $a_i$ if $a_i<0$, and 1 otherwise.

At the end of an episode (i.e. a predefined number of action steps that the agent is allowed to perform the task), the connection weights $w_{ij}$ are as follows:
\begin{equation}
w_{ij}^{\prime} = w_{ij} + \eta \cdot \Delta w_{ij}
\label{eq:npspEpisodicSynapticChange}
\end{equation}
\vspace{-0.5cm}
\begin{equation}
\Delta w_{ij} = NPSP(NAT_{ij},\theta)
\label{eq:npspSynapticChange}
\end{equation}
Finally, we scale the incoming connections in order to have a unit length:
\begin{equation}
w^{\prime}_{ij} = \frac{w^{\prime}_{ij}}{|| \boldsymbol{w^{\prime}_{i}} ||_2}  \label{eq:normEquation}
\end{equation}
This avoids increasing/decreasing the connection weights indefinitely, and also introduces synaptic competition.  

The \textit{eligibility traces} were proposed to trace the pairwise activations of pre- and post-synaptic neurons during an episode~\cite{gerstner2018eligibility}. Data structures inspired by the eligibility traces were previously employed to associate the pairwise neuron activations with reinforcement signals~\cite{yaman2019learning,izhikevich2007solving,soltoggio2013solving}. Shown in Table~\ref{tab:NATdataStructure}, we use \textit{neuron activation traces (NATs)} in each synapse to keep track of their activations (i.e. frequencies: $f_{00},f_{01},f_{10},f_{11}$) to be used in synaptic plasticity rules. We employ a threshold $\theta$ to convert the frequencies to binary representation. For instance, if a frequency value is lower than $\theta$, we assign 0, otherwise 1.

\begin{table}[H]
\begin{center}
\caption{The NAT data structure. For each connection $w_{ij}$, $NAT_{ij}$ stores the number of occurrences of each type of binary activation states of neuron pairs $i,j$.}\label{tab:NATdataStructure}
\begin{tabular}{|c|c|c|c|}
\hline
\multicolumn{4}{|c|}{$\boldsymbol{NAT_{ij}}$} \\ \cline{1-4}
$\boldsymbol{a_i=0,a_j=0}$             & $\boldsymbol{a_i=0,a_j=1}$            & $\boldsymbol{a_i=1,a_j=0}$           & $\boldsymbol{a_i=1,a_j=1}$ \\ \hline
$f_{00}$ & $f_{01}$ & $f_{10}$ & $f_{11}$\\ \hline
\end{tabular}
\end{center}
\end{table}

The goal in this case is to find how to perform synaptic changes based on the binary NAT values such that the network produces novel behaviors. Thus, as illustrated in Table~\ref{tab:NPSPNatrule}, we use a genetic algorithm (GA) to find weight updates ($x_1, x_2, \hdots, x_{16}$) for all possible states of 4 dimensional binary vectors. Each of these synaptic updates can be one of three values $\{-1,0,1\}$, that indicate increase, stable or decrease respectively (thus there are a total of $16^3$ possible plasticity rules).  

\begin{table}[H]
\begin{center}
\footnotesize
\caption{A list of binarized NATs states (based on a threshold $\theta$) are shown in a tabular form. The synaptic changes $x_1, x_2, \hdots, x_{16}$ are performed based on the NATs.}\label{tab:NPSPNatrule}
\begin{tabular}{|c|c|c|c|c|}
\hline
\multicolumn{4}{|c|}{$\boldsymbol{NAT}$}                                                                 & \multirow{2}{*}{$\boldsymbol{\Delta w}$} \\ \cline{1-4}
$\boldsymbol{a_i=0,a_j=0}$             & $\boldsymbol{a_i=0,a_j=1}$            & $\boldsymbol{a_i=1,a_j=0}$           & $\boldsymbol{a_i=1,a_j=1}$            &                                             \\ \hline\hline
$0$ & $0$ & $0$ & $0$ & $x_1$                 \\ \hline
$0$ & $0$ & $0$ & $1$ &   $x_2$                 \\ \hline
$\hdots$                     & $\hdots$                    & $\hdots$                     & $\hdots$                     & $\hdots$                \\ \hline
\multicolumn{1}{|c|}{1} & \multicolumn{1}{c|}{1} & \multicolumn{1}{c|}{1} & \multicolumn{1}{c|}{1} &             $x_{16}$       \\ \hline
\end{tabular}
\end{center}
\end{table}
%
%

The reason for using binary representations is to limit the search space. In addition, discrete rules (as shown in Table~\ref{tab:NPSPNatrule}) allow interpretability since they can be converted into a set of ``if-then'' statements. This may be more difficult when more complex functions (e.g., ANNs) are used to perform the synaptic changes.

\section{Experimental Setup} \label{sec:ExperimentalSetup}

In this section, we provide the details of our experimental setup. We designed deceptive maze environments and used the NPSP to produce novel behaviors to find a behavior that can achieve the goal. Since we do not use fitness values, we take random search and random walk algorithms as baseline. These tasks require complex actions to solve. Therefore, we use recurrent neural networks with various sizes. We discuss the details of the environments, agent architecture, genetic algorithm to evolve the NPSP rules and benchmark algorithms in following sections.

\subsection{Deceptive Maze Environments}

We perform experiments on environments that we refer to as deceptive maze (DM), because in these cases it is not straightforward to specify a fitness function to solve these tasks. Moreover, the use of simple fitness functions (such as the Euclidean distance to the goal) is usually deceiving to solve these problems since these functions are usually prone to get stuck in a local optimum and thus prevent finding good solutions~\cite{auerbach2016gaining,lehman2008exploiting}.

\begin{figure}[]
\begin{center}
\begin{subfigures}
\subfloat[DM11 door closed]{\includegraphics[width=0.48\columnwidth]{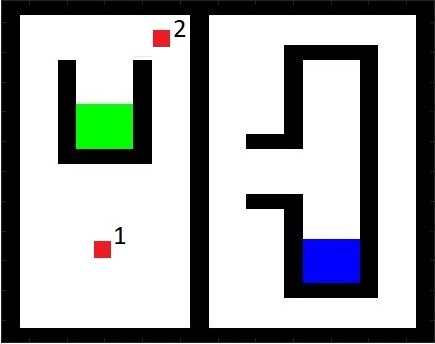} \label{fig:environment1closed}} \subfloat[DM12 door opened]{\includegraphics[width=0.48\columnwidth]{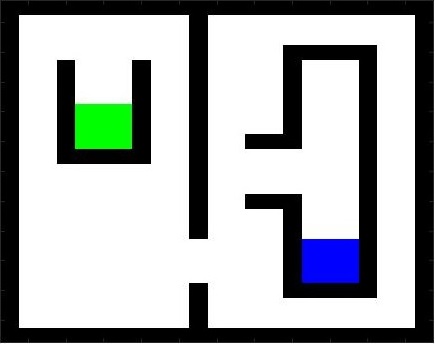} \label{fig:environment1open}}

\subfloat[DM21 door closed]{\includegraphics[width=0.48\columnwidth]{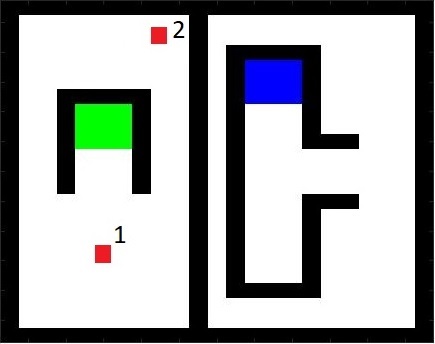} \label{fig:environment2closed}} \subfloat[DM22 door opened]{\includegraphics[width=0.48\columnwidth]{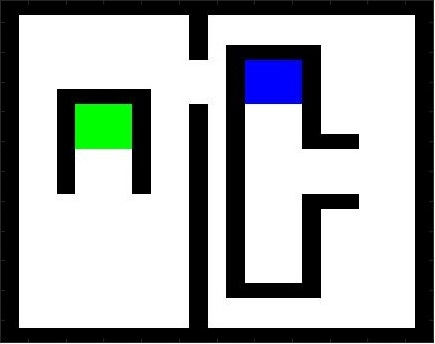} \label{fig:environment2open}}

\end{subfigures}
\end{center}

\caption{An illustration of two deceptive maze environments. Figures~\ref{fig:environment1closed} and \ref{fig:environment1open} are two versions of the first environment, and Figures~\ref{fig:environment2closed} and \ref{fig:environment2open} are two versions of the second environment. The only difference between the two versions of the same environment is an opening on the middle wall that allows agents to travel from the left room to the right. Labels ``1'' and ``2'' show two independent starting positions of the agent.} \label{fig:deceptiveMaze}
\end{figure}

Visual illustrations of the DM environments are shown in Figure~\ref{fig:deceptiveMaze}. The environments consist of $23 \times 23$ cells. Each cell can be occupied by one of five possibilities: \textit{empty, wall, goal, button, agent}, color-coded in white, black, blue, green, red respectively. The starting position of the agent is illustrated in red. There are two starting positions of the agent labelled as ``1'' and ``2''. These starting positions are tested separately.

Figures~\ref{fig:environment1closed} and \ref{fig:environment1open} show two versions of the same environment, that we refer to as DM11 and DM12, and Figures~\ref{fig:environment2closed} and \ref{fig:environment2open} show two versions of the same environment we refer as DM21 and DM22. The difference between two versions of the same environment is that there is an opening (door) in the middle of the wall to allow the agent to travel between rooms when it is open.

Starting from one of the starting positions, the behavior that solves the task involves first going to the button area (in green) and perform a ``press button" action. In this case, the door in the middle of the wall opens. The agent is then required to pass through this opening and reach the goal position (in blue).

\subsection{Agent Architecture}

\begin{figure}[]
\begin{center}
\begin{subfigures}
\subfloat[]{\includegraphics[width=0.55\columnwidth]{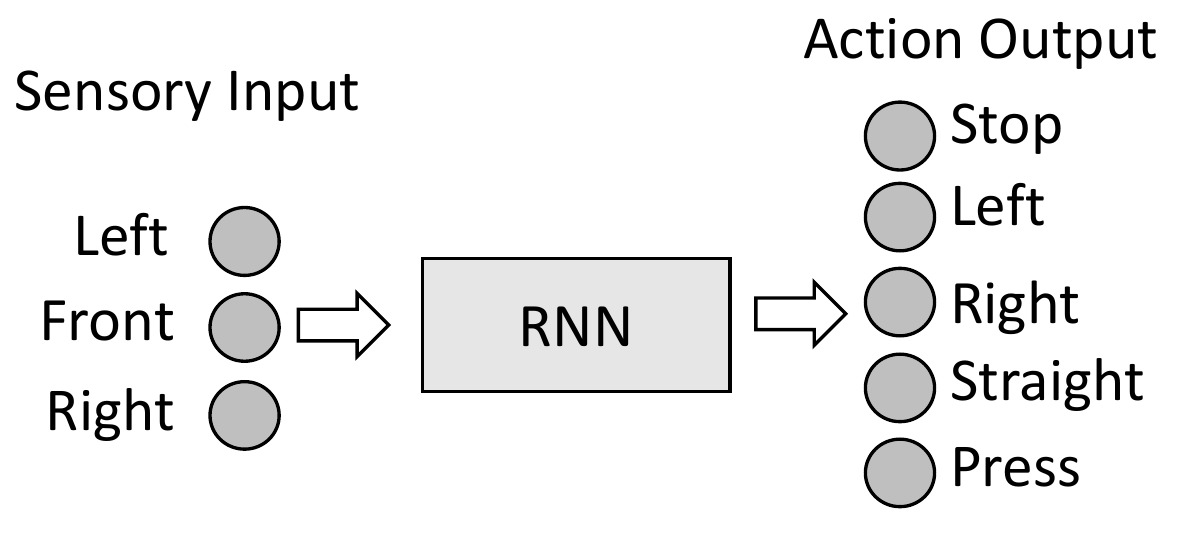}\label{fig:agent}} \subfloat[]{\includegraphics[width=0.4\columnwidth]{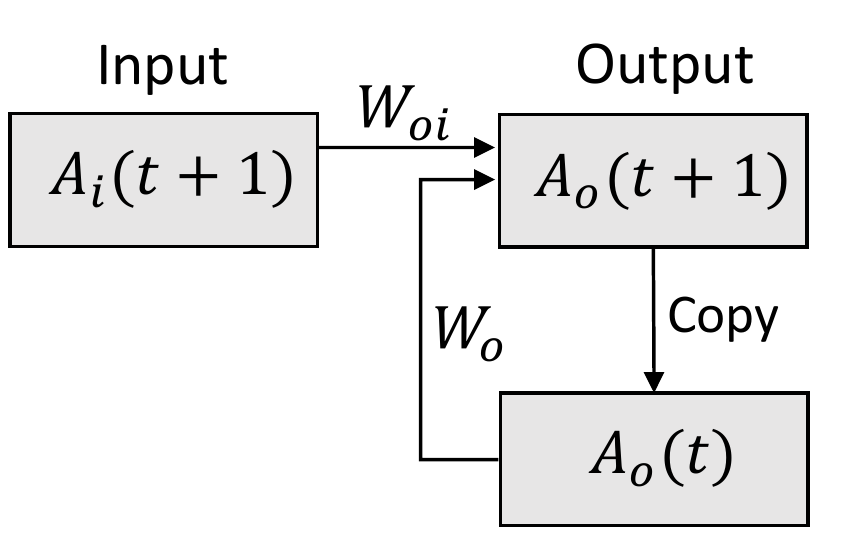}\label{fig:rnnArchitecture}}

\subfloat[]{\includegraphics[width=0.6\columnwidth]{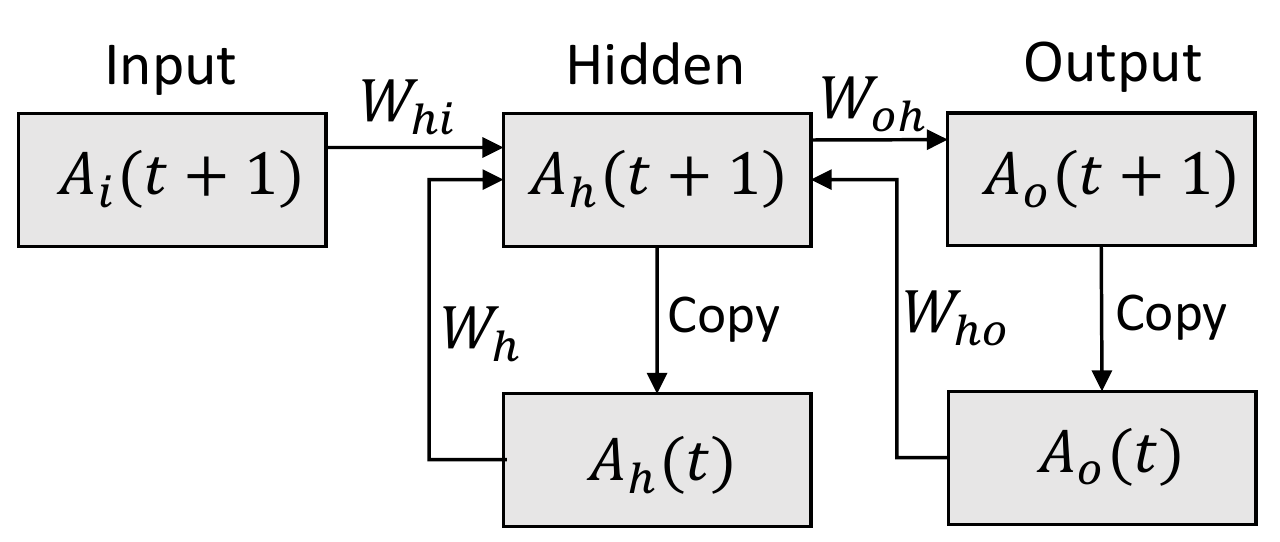}\label{fig:rnnArchitectureHidden}}
\end{subfigures}
\end{center}
\caption{(a): The sensory inputs and action outputs of the recurrent neural networks that are used to control the agents; (b) and (c): The architectures of the network without and with a hidden layer respectively.} \label{fig:agentArchitecture}
\end{figure}

An illustration of the architecture of the agents used for the deceptive maze tasks is given in Figure~\ref{fig:agentArchitecture}. In each action step, the agent can take the nearest right, front, and left cells as inputs and perform one of the actions as: stop, left, right, straight, press. Each input sensor can sense if there is a wall or not (represented as 0 and 1 respectively). The door opens when the press action is performed only if the agent is within the button area (green). Multiple press actions while the agent is within the button area do not have any effect. 


Illustrated in Figures~\ref{fig:rnnArchitecture} and \ref{fig:rnnArchitectureHidden}, we use two types of RNNs (without and with a hidden layer) to control the agents. The network shown in Figure~\ref{fig:rnnArchitecture} consists of 40 connection parameters as: $W_{oi}: (3+1) \times 5 = 20$ ($+1$ refers to the bias), $W_o:(5-1)\times 5 = 20$ (except self-node connections), and the network shown in Figure~\ref{fig:rnnArchitectureHidden} consists of 15 hidden neurons and 4 sets of connections between the layers as: $W_{hi}:(3+1) \times 15 = 60$, $W_h:(15-1) \times 15 = 210$ (except self-node connections), $W_{oh}:15 \times 5 = 75$, $W_{ho}:5 \times 15 = 75$. Thus, the network has in total 420 parameters. We used the network without the hidden layer and the network with 15 hidden neurons to limit the computation during the evaluation process. We further tested evolved NPSP rules on networks with 30 and 50 hidden neurons. These networks have 1290 and 3150 parameters respectively.

As for the networks without a hidden layer, the activations of the output layer are computed as: 
\begin{equation}
\boldsymbol{A}_o(t+1) = \psi 
\Big( 
\boldsymbol{W}_{oh} \cdot \boldsymbol{A}_i(t) + \alpha_o \cdot \boldsymbol{W}_{oh} \cdot \boldsymbol{A}_o(t)
\Big)
\label{eq:outputActivationWithoutHidden}
\end{equation}

In the case of networks with a hidden layer, the activations of the hidden and output layers are computed as:
\begin{equation}
\begin{split}
\boldsymbol{A}_h(t+1) = & \psi 
\Big( \boldsymbol{W}_{hi} \cdot \boldsymbol{A}_i(t+1) + \alpha_h \cdot \boldsymbol{W}_{h} \cdot \boldsymbol{A}_h(t) \\ & + \alpha_o \cdot \boldsymbol{W}_{ho} \cdot \boldsymbol{A}_o(t) \Big)
\end{split}
\label{eq:hiddenActivation}
\end{equation}
\begin{equation}
\boldsymbol{A}_o(t+1) = \psi 
\Big( 
\boldsymbol{W}_{oh} \cdot \boldsymbol{A}_h(t+1)
\Big)
\label{eq:outputActivation}
\end{equation}
where the parameters $\alpha_h$ and $\alpha_o$ are added to scale the recurrent and feedback connections. Parameter $t$ denotes the time step.

\subsection{Genetic Algorithm}
The NPSP rules consist of discrete and continuous parts. A standard GA was used to evolve the discrete parts of the NPSP rules. The discrete parts of the genotypes consist of 16 genes, initialized randomly from $\{-1,0,1\}$. The continuous parts of the genotypes are initialized randomly from these ranges: $\eta\in[0,1],\theta \in [0,1],\alpha_h\in [0,1],\alpha_o\in [0,1]$. Thus, the genotype of the individuals is represented by a 20-dimensional discrete/real-valued vector (19-dimensional in the case without a hidden layer). 

We evaluate each NPSP rule on two environments, illustrated in Figure \ref{fig:deceptiveMaze}, for two starting positions and three trials each. Thus, in total, we perform $N_{trials} = 12$ independent trials. Each of these trials consists of $N_{episodes}=500$ episodes of learning process, where each episode consists of 250 action steps to reach the goal from the starting position.

The fitness value of an individual NPSP rule is computed as:
\begin{equation}
fitness = \frac{1}{N_{trials}} \sum_{n=1}^{N_{trials}} \frac{\big|unique(\boldsymbol{B_n})\big|}{N_{episodes}}
\label{eq:fitnessNPSP}
\end{equation}
which is based on the average number of novel behaviors the NPSP rule produces. To calculate that, we abstract and record the behavior of an agent during each episode and append it to the behavior set $\boldsymbol{B}$, and find the average number of novel (unique) behaviors per trial. 
\begin{figure}[]
\begin{center}
\includegraphics[width = 0.5\columnwidth]{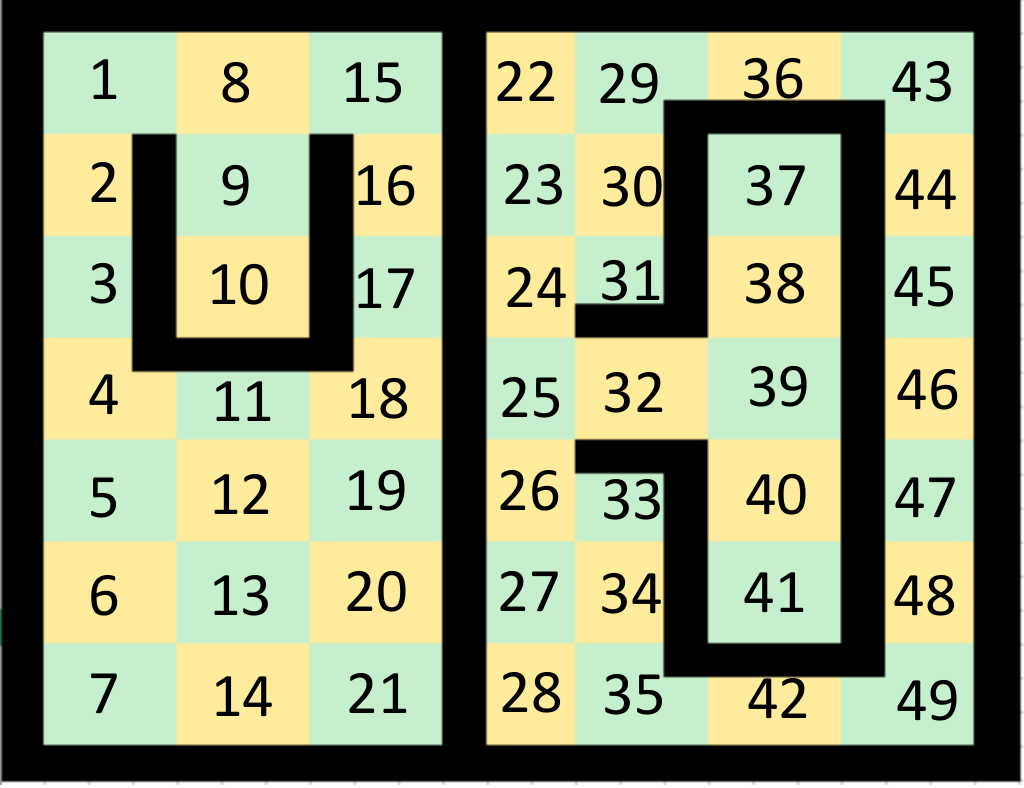}
\caption{An example illustration of the environment representation that is used to abstract the behavior of the agents.}
\label{fig:EnvironmentBehavior}
\end{center}
\end{figure}

The behavior abstraction is performed as follows. The environment is divided in $3 \times 3$ squares, as shown in Figure~\ref{fig:EnvironmentBehavior}, and each square is given two unique identifiers (ids) (e.g. ``1'' and ``1*'') to distinguish between two states of the agent: ``located in the square'' and ``located in the square and pressing the button''. Inspired by Pugh~\textit{et al.,}~\cite{pugh2015confronting}, we abstract the behavior of an agent by recording its trajectory based on the locations visited, and save it as a sequence of ids in a string form. For instance, one example string could be:``13-13*-12-11-4-3-2-1-8-9-10-10*''. This string means that the agent started from square 13, next performed a press button action while it was in square 13, next passed through of a sequence of squares 12, 11, 4, $\dots$, 10, then finally performed a press button action while it was on square 10. We do not repeat the square id if the agent is staying in the same square for more than one time step. We refer to a string like this as a behavior. We collect the behavior in each episode and find how many novel behaviors the NPSP rule is able to produce during one trial (500 learning episodes). This is achieved by finding how many novel sequences were generated. Thus, we aim to maximize the number of novel behaviors produced, in the attempt to find the behavior that solves the task.

\begin{figure}[]
\begin{center}
\begin{subfigures}
\subfloat[]{\includegraphics[width=0.49\columnwidth]{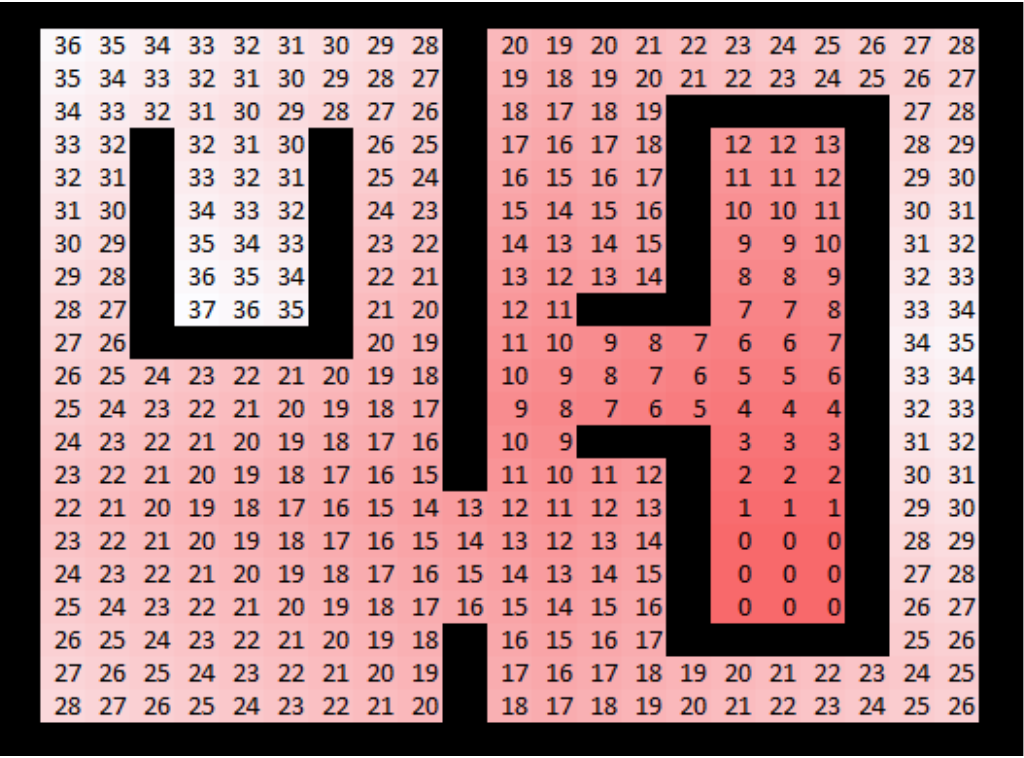}\label{fig:Environment1DistanceMatrix}}\hfill \subfloat[]{\includegraphics[width=0.49\columnwidth]{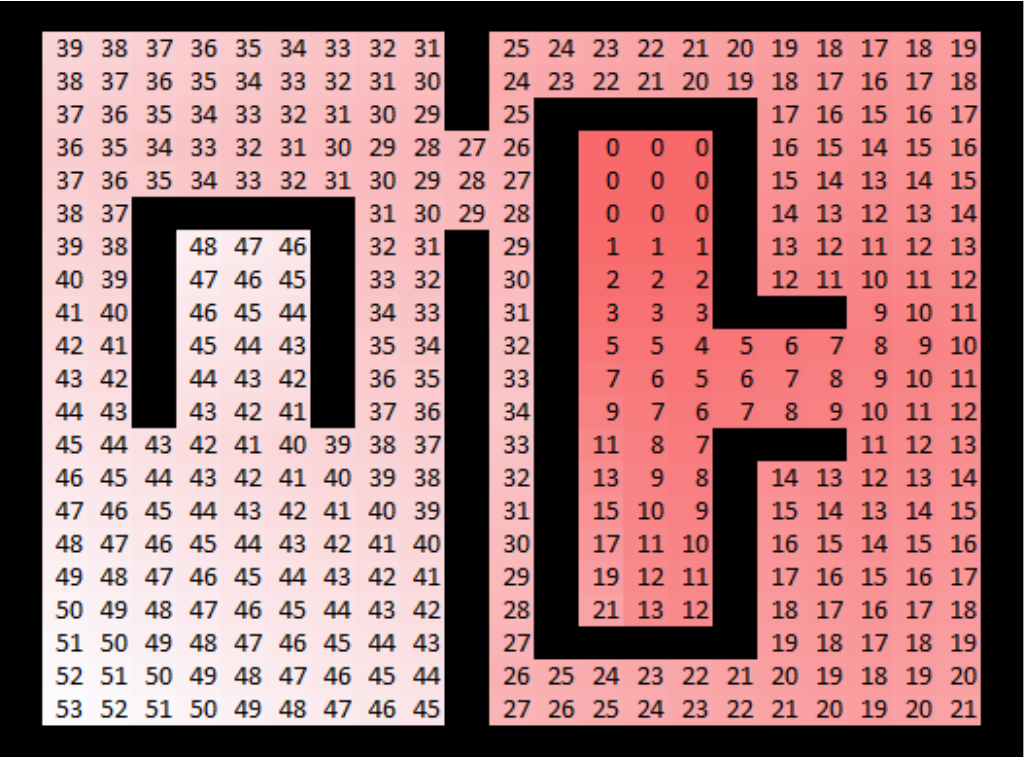}\label{fig:Environment2DistanceMatrix}}
\end{subfigures}
\end{center}
\caption{The distances of each cell to the goal position in each environment are shown as heatmaps where the intensity of red color indicates lower distance.} \label{fig:EnvironmentDistanceMatrices}
\end{figure}

The distances of each cell to the goal position in the environments are measured as shown in Figure~\ref{fig:EnvironmentDistanceMatrices}. During each episode, the closest distance $min(d_e(XY(Agent),XY(g)))$ to the goal position $XY(g)$ is also recorded.

The comparison is performed based on two performance measures: ``novelty'' and ``distance''. The latter is the average of the smallest distances to the goals that an agent achieved during the episodes. Both these measures are scaled within a certain range to  make it easy to perform comparisons between the results of different runs related to different starting points and different environments. Thus, the novelty measure is divided by the number of episodes, to scale it between 0 and 1. The higher the novelty score of an agent, the more novel behaviors it has produced. The distance measure is adjusted depending on whether the agent manages to pass through the door to the second room where the goal is located. If the agent is not able to pass to the second room, its distance measure is updated as:
\begin{equation}
dist_{agent} = 1 +\frac{min(d_e(XY(Agent),XY(g)))}{maxDist}
\label{eq:npspFitnessMeasure1}
\end{equation}
Otherwise (if the agent manages to go to the second room where the goal is located), its distance measure is updated as:
\begin{equation}
dist_{agent} = \frac{min(d_e(XY(Agent),XY(g)))}{maxDistSecondRoom}
\label{eq:npspFitnessMeasure2}
\end{equation}
where $maxDist$ and $maxDistSecondRoom$ are constant values indicating the maximum distance to the goal, and the maximum distance to the goal in the second room. Thus, the updated distance measure is between 0 and 2. If it is greater than 1, it means that the agent was not able to pass to the second room; and if it is smaller than 1, it means that the agent managed to pass to the second room. Overall, its value indicates the distance to the goal position, the smaller means the closer.

We use a population size of 14 and employ a \textit{roulette wheel selection} operator with an elite number of four. We use a \textit{1-point crossover} operator with a probability of 0.5 and a custom \textit{mutation} operator which re-samples each discrete dimension of the genotype with a probability of 0.15 and performs a Gaussian perturbation with zero mean and 0.1 standard deviation for the continuous parameters. We run the evolutionary process for 100 generations. In each generation of the evolutionary process, we store the NPSP rules that produced the largest number of novel behaviors, and the NPSP rules that achieved the minimum distance to the goal positions.

\subsection{Benchmark Algorithms}

We use two analogous algorithms, Random Search (RS) and Random Walk (RW), to perform comparisons with the NPSP rules. The RS and RW algorithms use a single solution to perform synaptic changes after every episode. However, they perform synaptic changes by random initialization and perturbation, respectively, without using any domain knowledge on the neuron activation as it is introduced with the NPSP rules.

\begin{figure}[H]
\begin{center}
\begin{subfigures}
\subfloat[]{\includegraphics[width=0.40\columnwidth]{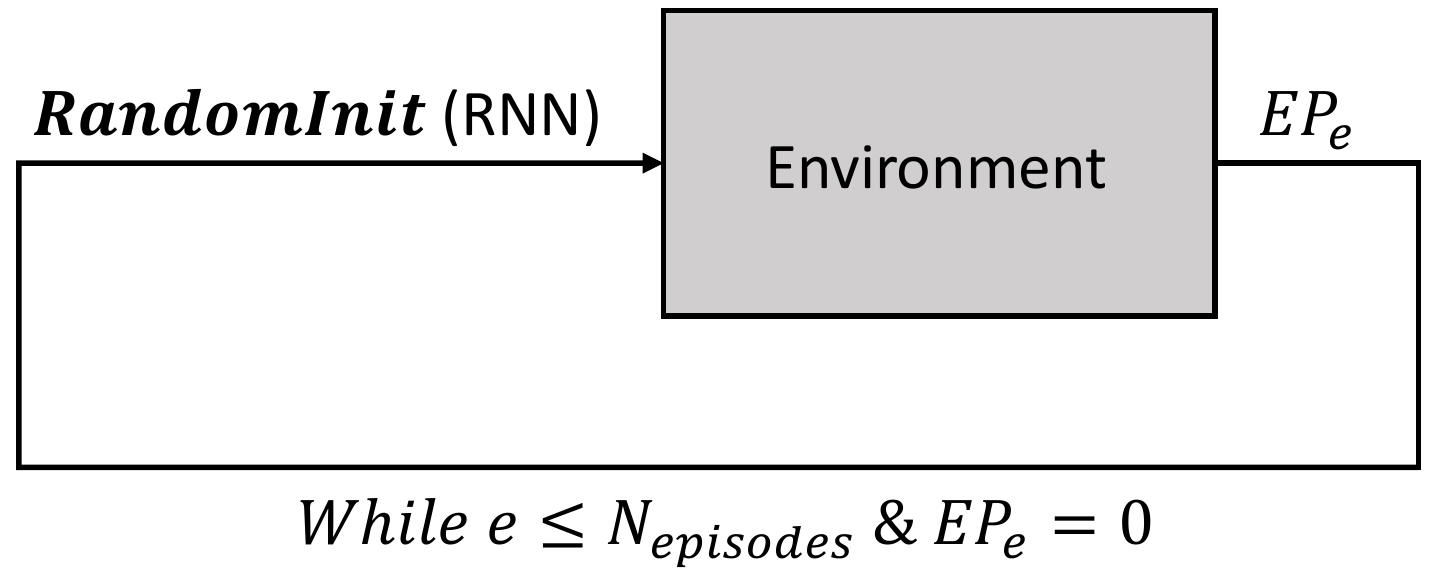}\label{fig:randomSearchNPSP}} \subfloat[]{\includegraphics[width=0.6\columnwidth]{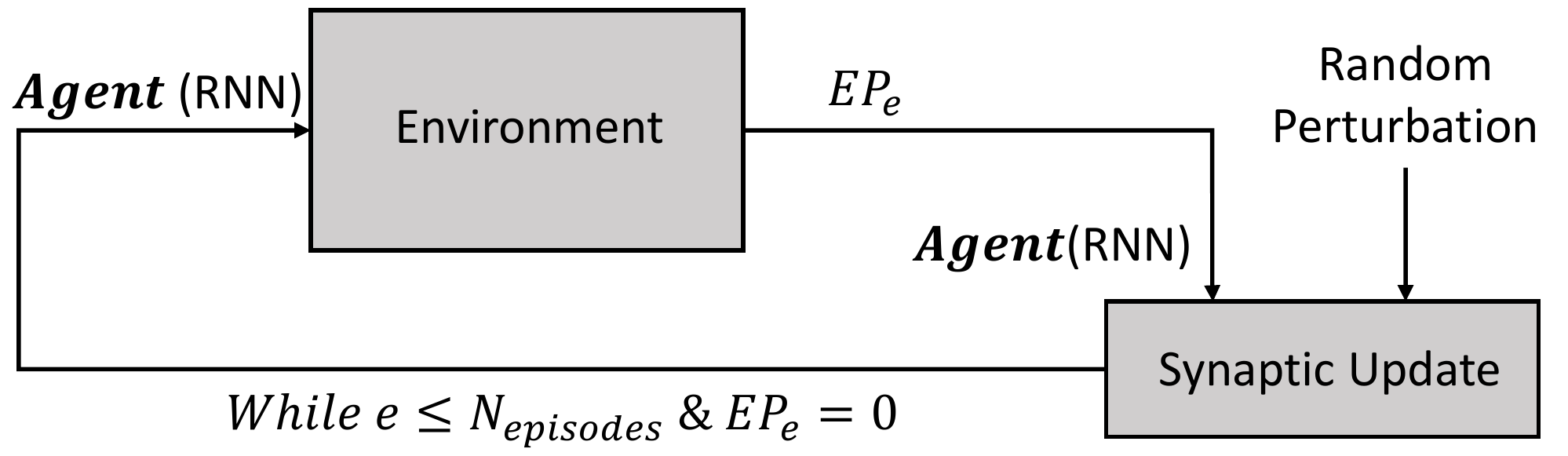}\label{fig:randomWalkNPSP}} 

\subfloat[]{\includegraphics[width=0.7\columnwidth]{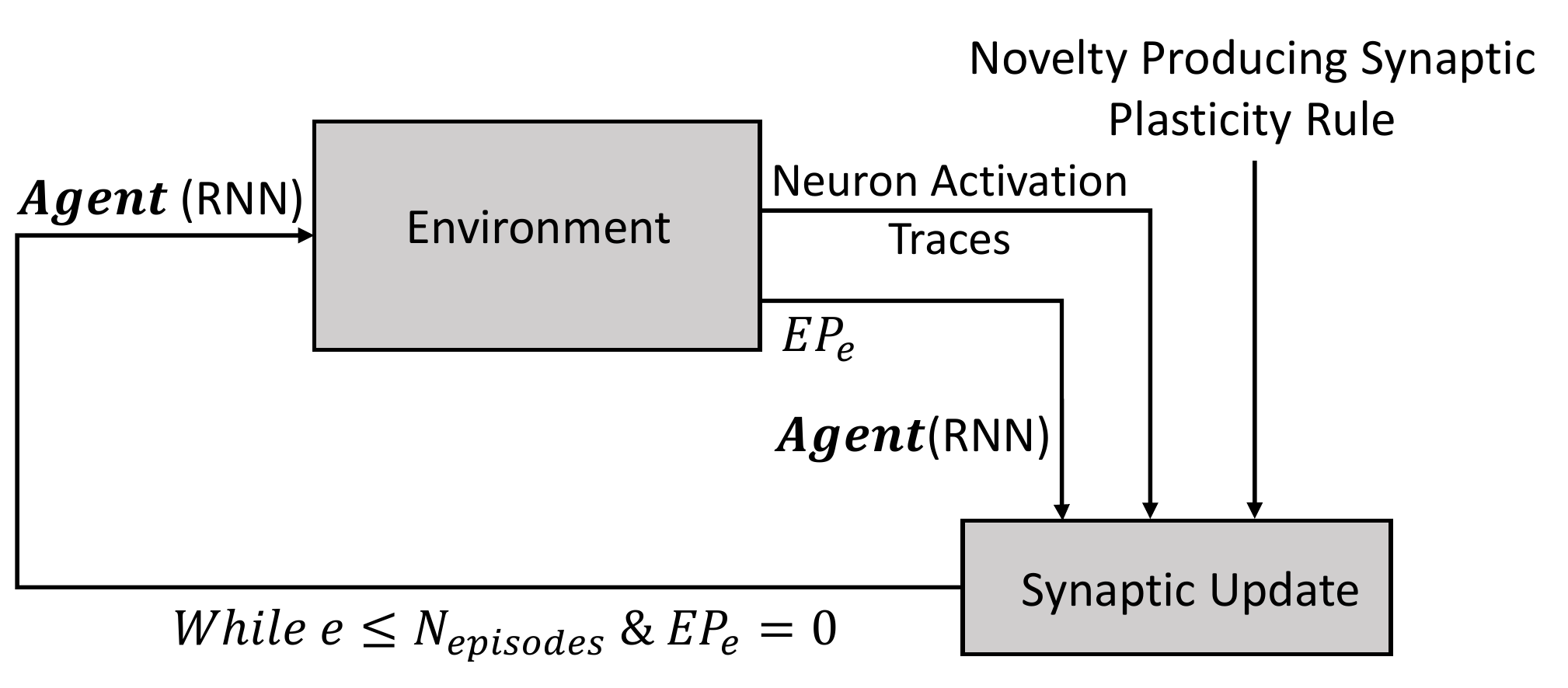}\label{fig:noveltyProducingSP}}

\end{subfigures}
\end{center}

\caption{The learning process of the RNNs that controls the agents using: (a) random search, (b) random walk, (c) novelty producing synaptic plasticity.} \label{fig:frameworkNPSP}
\end{figure}

Figure~\ref{fig:frameworkNPSP} shows the learning processes with RS, RW and NPSP. All algorithms start with randomly initialized RNNs which are used to control the agent within the environment for an episode. At the end of an episode, we obtain the episodic performance as $EP = 1$ or $EP = 0$, which indicates that either the task is solved or not. If the task is not solved, we perform synaptic changes and test again the agent on the task. This process continues for a certain number of episodes $N_{episodes}$, or until the task is solved. In the case of RS, after each episode the network is re-initialized. In the case of RW, the weights of the network are perturbed by Gaussian perturbation with standard deviation $\sigma$ as: $w_{ij} = w_{ij} + \mathcal{N} (0,\sigma )$. Thus, the RS performs random search in the search space, whereas the RW performs a random search within the neighboring networks of the initial network. In the case of the NPSP, we use the evolved rules to perform perturbations.


\section{Experimental Results}\label{sec:ExperimentalResults}

In this section, we present the results of the agents trained using RS, RW and NPSP rules. The comparisons between the results of the algorithms are performed based on the novelty and distance measures that are explained in Section~\ref{sec:ExperimentalSetup}.

Table~\ref{tab:compResults} shows the median of the novelty and distance measures of the agents trained by RS, RW and evolved NPSP rules. The columns labelled as ``Goal'' and ``Second Room'' report the number of times the agents were able to achieve the goal and enter into the second room respectively. For all algorithms, the learning process is set to 500 episodes and 12 trials in total (3 trials for 2 starting positions, for 2 environments). 

The rows labelled as RS0H, RW0H and NPSP0H show the results of the algorithms on the RNN models without a hidden layer. We observe that RS0H produces more novel behaviors relative to RW0H. This could be expected since RS0H randomly samples from the search space after each episode, whereas RW0H performs iterative perturbations on randomly initialized solutions, thus it performs the search more locally. Consequently, RS0H leads to lower distance measure. 

\begin{table}[b]
\centering
\small
\caption{The median of the novelty and distance measures of agents trained by RS, RW and the evolved best performing NPSP rule.}\label{tab:compResults}
\begin{tabular}{|c|c|c|c|c|}
\hline
\textbf{Algorithm}  & \textbf{Novelty} & \textbf{Distance} & \textbf{Goal} & \textbf{Second Room}  \\ \hline
\textbf{RS0H}                  & 0.095            & 1.3974            & 0             & 0                                     \\ \hline
\cellcolor[HTML]{EFEFEF}\textbf{RW0H}                  &\cellcolor[HTML]{EFEFEF}0.018            & \cellcolor[HTML]{EFEFEF}1.5032            & \cellcolor[HTML]{EFEFEF}0             & \cellcolor[HTML]{EFEFEF}0                                   \\ \hline
\textbf{NPSP0H}             & 0.3550   & 0.6786            & 2            & 7                \\ \hline\hline

\textbf{RS15H}                     & 0.2583                & 1.3302            & 0             & 2                                \\ \hline
\cellcolor[HTML]{EFEFEF}\textbf{RW15H}                     & \cellcolor[HTML]{EFEFEF}0.2228                    &\cellcolor[HTML]{EFEFEF}1.4533            & \cellcolor[HTML]{EFEFEF}0             & \cellcolor[HTML]{EFEFEF}0                                    \\ \hline
\textbf{NPSP15H}                      & 0.4110                          & 0.8393            & 0             &8                                \\ \hline\hline

\textbf{RS30H}                     & 0.4328                & 1.2856            & 0             & 3                                   \\ \hline
\cellcolor[HTML]{EFEFEF}\textbf{RW30H}                    & \cellcolor[HTML]{EFEFEF}0.3022                   & \cellcolor[HTML]{EFEFEF}1.2944         & \cellcolor[HTML]{EFEFEF}0             & \cellcolor[HTML]{EFEFEF}3                                  \\ \hline
\textbf{NPSP30H}                       &  0.8400                           & 0.8571            & 1             & 7                        \\ \hline\hline

\textbf{RS50H}                      & 0.6300    & 0.8920            & 0             & 7                                  \\ \hline
\cellcolor[HTML]{EFEFEF}\textbf{RW50H}                      & \cellcolor[HTML]{EFEFEF}0.5072        & \cellcolor[HTML]{EFEFEF}1.1606            & \cellcolor[HTML]{EFEFEF}0             & \cellcolor[HTML]{EFEFEF}3 \\ \hline
\textbf{NPSP50H}                      & 1.0000    & 0.5179            & 1             & 8                                 \\ \hline


\end{tabular}

\end{table}

NPSP0H was selected after six independent evolutionary runs because it produced the highest number of novel behaviors. The agent trained with NPSP0H was able to produce about 177 ($0.3550\times 500$) novel behaviors on average, and was able to enter into the second room in 7 out of 12 trials. 

The rest of the rows shows the comparison results of the networks with hidden layers. Similarly, we performed two independent evolutionary runs on RNNs with 15 hidden neurons and optimized the NPSP rules. We then selected the best NPSP rule, that is NPSP15H, and tested on the RNNs with 15, 30 and 50 hidden neurons. The results are labelled as NPSP15H, NPSP30H and NPSP50H respectively.

We observe quite interestingly that the algorithms produce larger number of novel behaviors when the number of hidden neurons are increased. For instance, RS15H produces about 129 ($0.2583 \times 500$) novel behaviors and RS50H  produces about 315  ($0.63 \times 500$) novel behaviors. On the other hand, the NPSP rule was able to produce much more novel behaviors compared to RS and RW for all sizes of the networks. For instance, NPSP50H was able to produce 500 novel behaviors in 500 episodes and yielded the lowest score for the median of distance to the goal in 12 trials (it also reached the second room in 8 trials). 

We noticed that NPSP0H was able to produce competitive results in terms of distance even thought it was not able to produce more novelty than the cases with hidden neurons. This may be due to the ``granularity'' of the behaviors produced. We would expect the RNNs with hidden layers (especially larger ones) to produce behaviors that are more complicated and detailed due to the large number of parameters that could affect the production of sequences of behaviors. On the other hand, we expect the RNNs without a hidden layer to produce more high level behavior patterns. This can explain why the smaller sized networks (i.e. without hidden layer) could produce less novel behaviors, and yet be successful in finding the behaviors that can get closer to the goal. They can produce high level and less complex behaviors (i.e. bouncing from the walls and following the walls) that may explore the environment. We have recorded several behaviors generated by the NPSP rules on RNN models with and without a hidden layer. Moreover, small changes in weights may lead to smaller behavioral differences relative to the small changes in larger networks, thus, as expected, we observe that a large number of novel behaviors is produced by larger networks, even though the same NPSP rule is used. We recorded a video, available online\footnote{A video recording of behaviors found by the NPSP rules using the RNN models with and without a hidden layer, accessible online at: http://bit.ly/2H4IOp5.}, to illustrate a visual comparison of the successful agent behaviors found by the NPSP rules using the RNN models with and without a hidden layer.

\begin{figure}[]
\begin{center}
\begin{subfigures}
\subfloat[Novelty Trend]{\includegraphics[width=0.49\columnwidth]{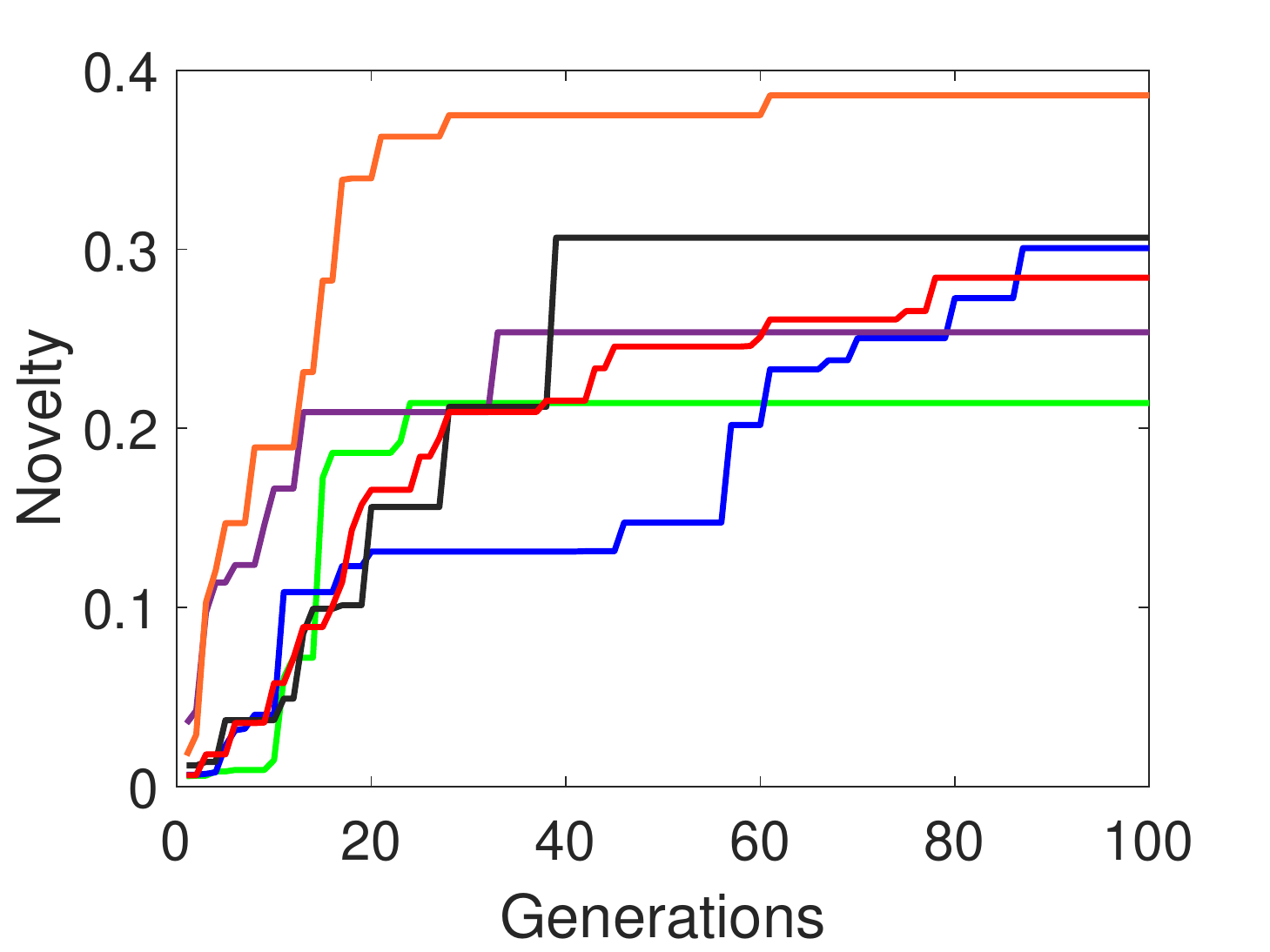}\label{fig:noveltyTrend}}\hfill \subfloat[Distance Trend]{\includegraphics[width=0.49\columnwidth]{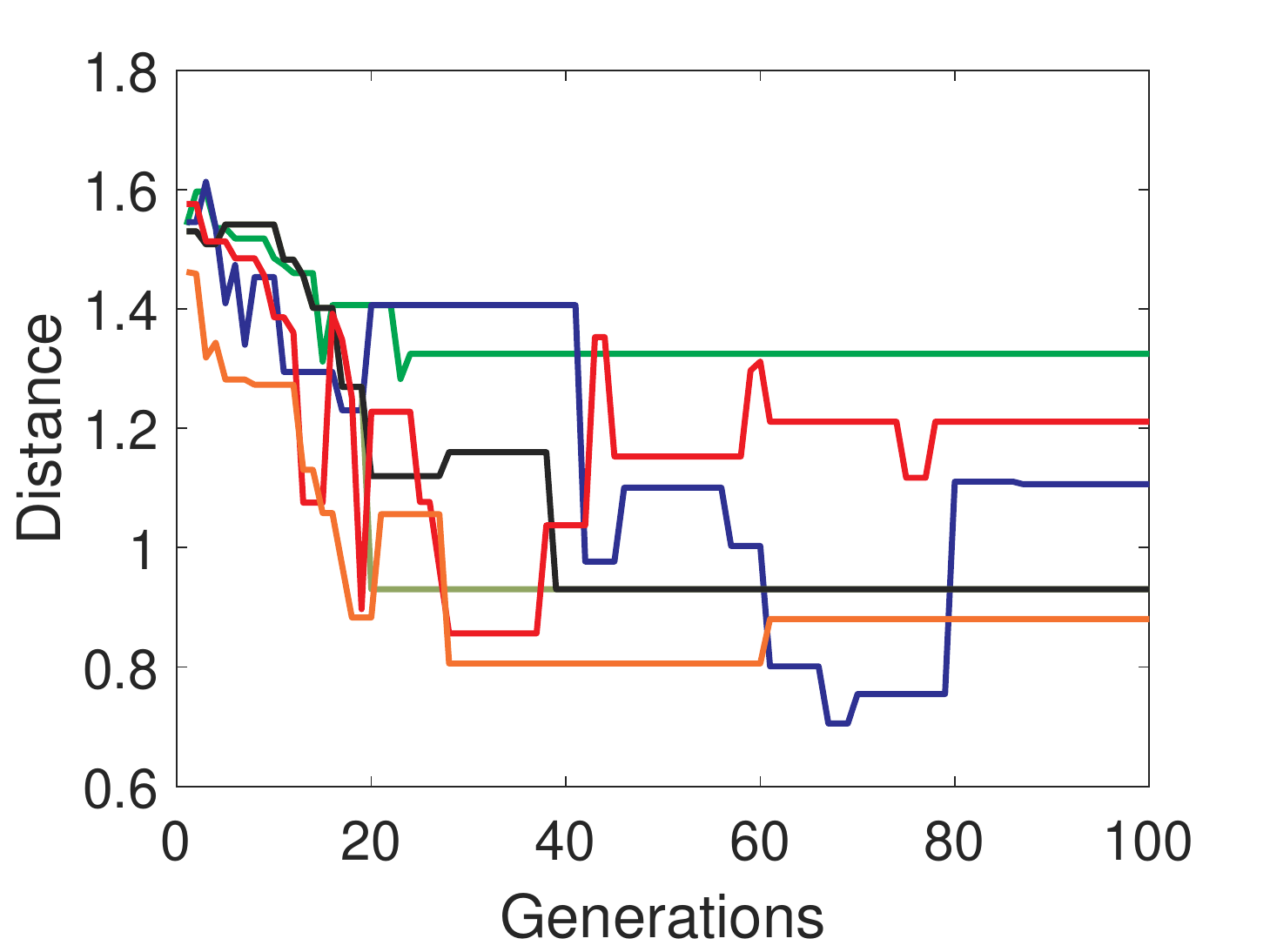}\label{fig:distanceTrend}}
\end{subfigures}
\end{center}
\caption{The novelty and distance trends of the NPSP rules during 6 independent evolutionary runs. } \label{fig:noveltyAndDistanceTrends}
\end{figure}

In Figures~\ref{fig:noveltyTrend} and \ref{fig:distanceTrend}, we illustrate the novelty and distance trends of six independent evolutionary runs of the NPSP rules optimized using the RNN model without a hidden layer. Since the NPSP rules were selected based on their novelty, their distance trends are not decreasing at all times. Thus, some rules showed better distance but had lower novelty score.


\begin{figure}[]
\begin{center}
\begin{subfigures}
\subfloat[DM1 Distance Measure]{\includegraphics[width=0.43\columnwidth]{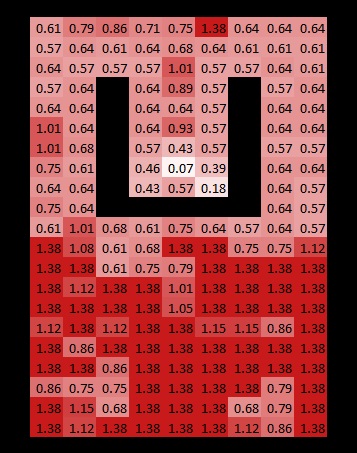}\label{fig:env1DistMeasur}}\hfill \subfloat[DM1 Novelty Measure]{\includegraphics[width=0.43\columnwidth]{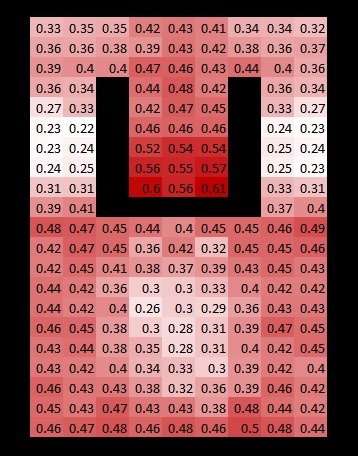}\label{fig:env1NovMeasur}}\\
\subfloat[DM2 Distance Measure]{\includegraphics[width=0.43\columnwidth]{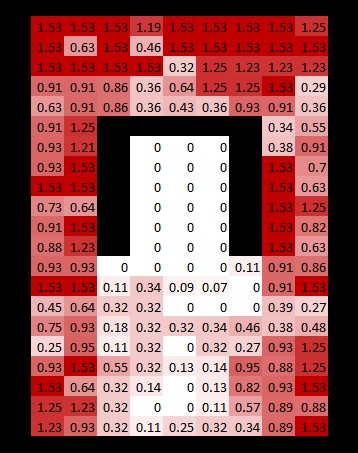}\label{fig:env2DistMeasur}}\hfill \subfloat[DM2 Novelty Measure]{\includegraphics[width=0.43\columnwidth]{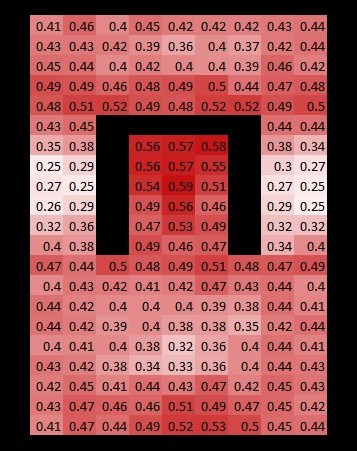}\label{fig:env2NovMeasur}}
\end{subfigures}
\end{center}
\caption{The median of the novelty and distance measures of 12 independent trials of the agents trained by NPSP0H. The value in each cell indicates the result when it is set as the starting point. Only the first rooms of the environments are shown since the agents can only start from there. The intensity of the colour indicates the magnitude of the value in each cell.} \label{fig:environmentMeasureDistribution}
\end{figure}

We further assess the performance of NPSP0H with respect to different starting positions. For that, we assigned each cell in the first room of DM1 and DM2 as the starting point of the agent and used NPSP0H to train the agent for 12 independent trials (each starting from a randomly initialized RNN configuration). Note that the NPSP rules were evolved based on two selected starting points but in this case they are tested on all locations, which may give some insights into the generalizability of their performance. The results are shown in Figure~\ref{fig:environmentMeasureDistribution} where we show the median of the distance and novelty measures in each cell when it is used as the starting point. We color-coded the figures based on the magnitude of the values in each cell where the intensity of red indicates higher values.

Based on the distance measures shown in Figures~\ref{fig:env1DistMeasur} and \ref{fig:env2DistMeasur}, we observe that the agents starting close the wall and behind the obstacle do not seem to get closer to the goal position. Correspondingly, Figures~\ref{fig:env1NovMeasur} and \ref{fig:env2NovMeasur} show lower novelty measures in similar areas. On the other hand, the agents that start from the middle area, and locations facing or within the button area, are capable of getting closer to the goal and also have higher novelty measure. 

Overall, the agents started from 172 cells in DM1 and DM2. The median result was below 1 (which means that the agents are able to access the second room) in 95 and 120 out of 172 starting points (55.2\% and 69.7\%) in DM1 and DM2 respectively. This shows that the agents in DM2 were more successful in getting closer to the goal. Therefore, this may indicate that the first environment is more difficult to solve, and/or NPSP0H may have an environmental bias towards DM2.

\begin{figure}[]
\begin{center}
\begin{subfigures}
\subfloat[ENV1 door closed]{\includegraphics[width=0.5\columnwidth]{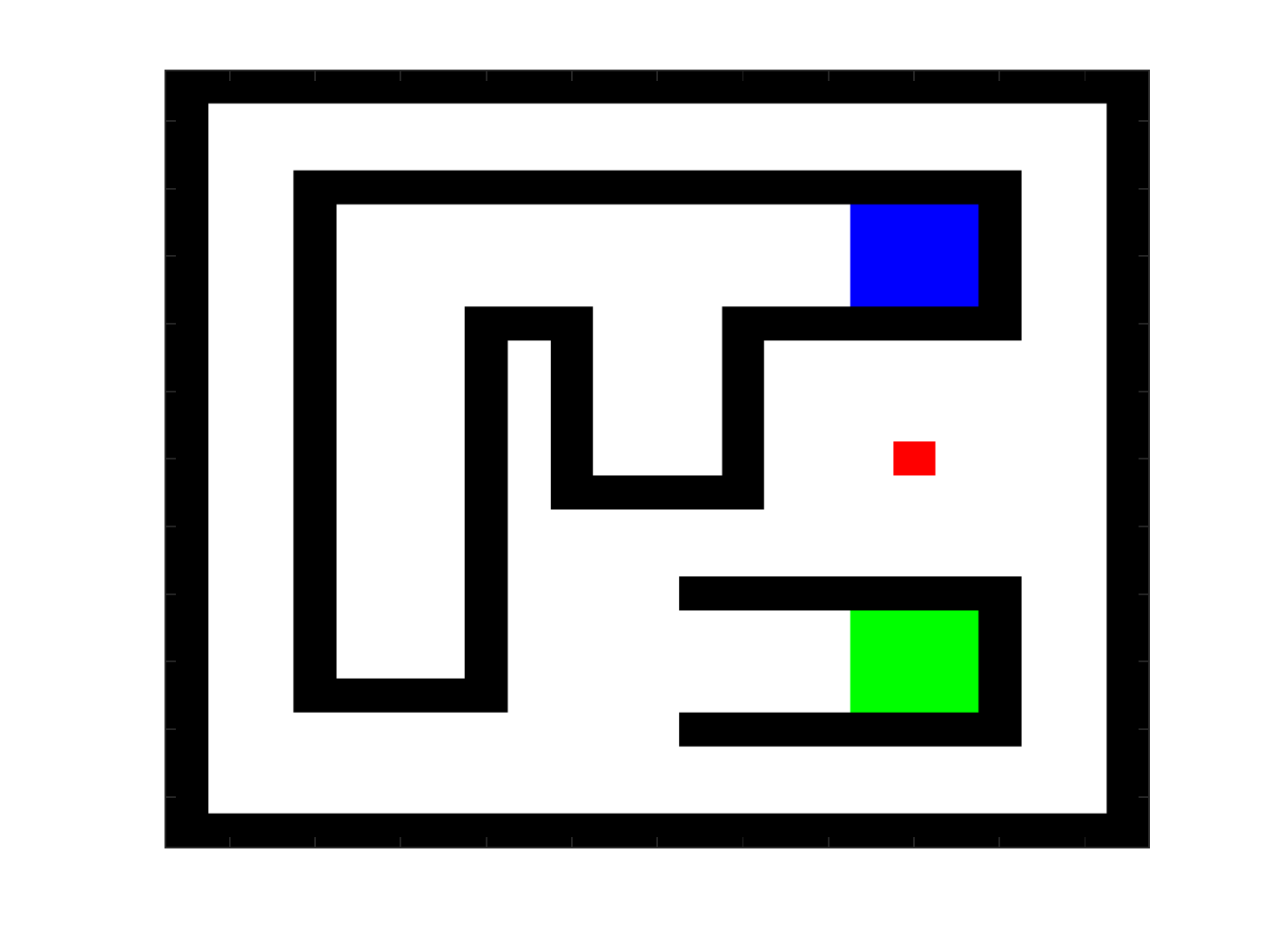} \label{fig:varEnv11}} \subfloat[ENV1 door opened]{\includegraphics[width=0.5\columnwidth]{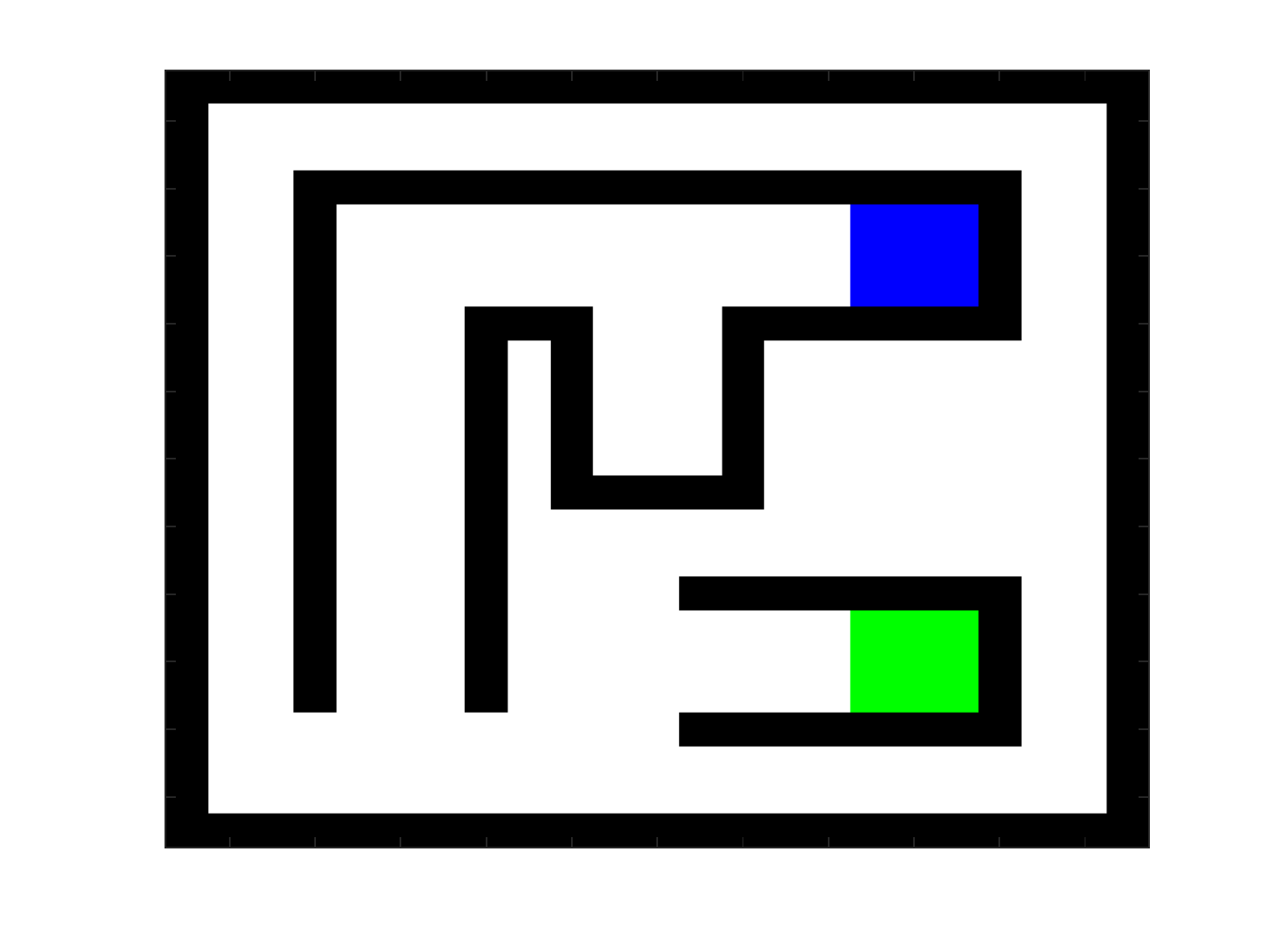} \label{fig:varEnv12}}

\subfloat[ENV2 door closed]{\includegraphics[width=0.5\columnwidth]{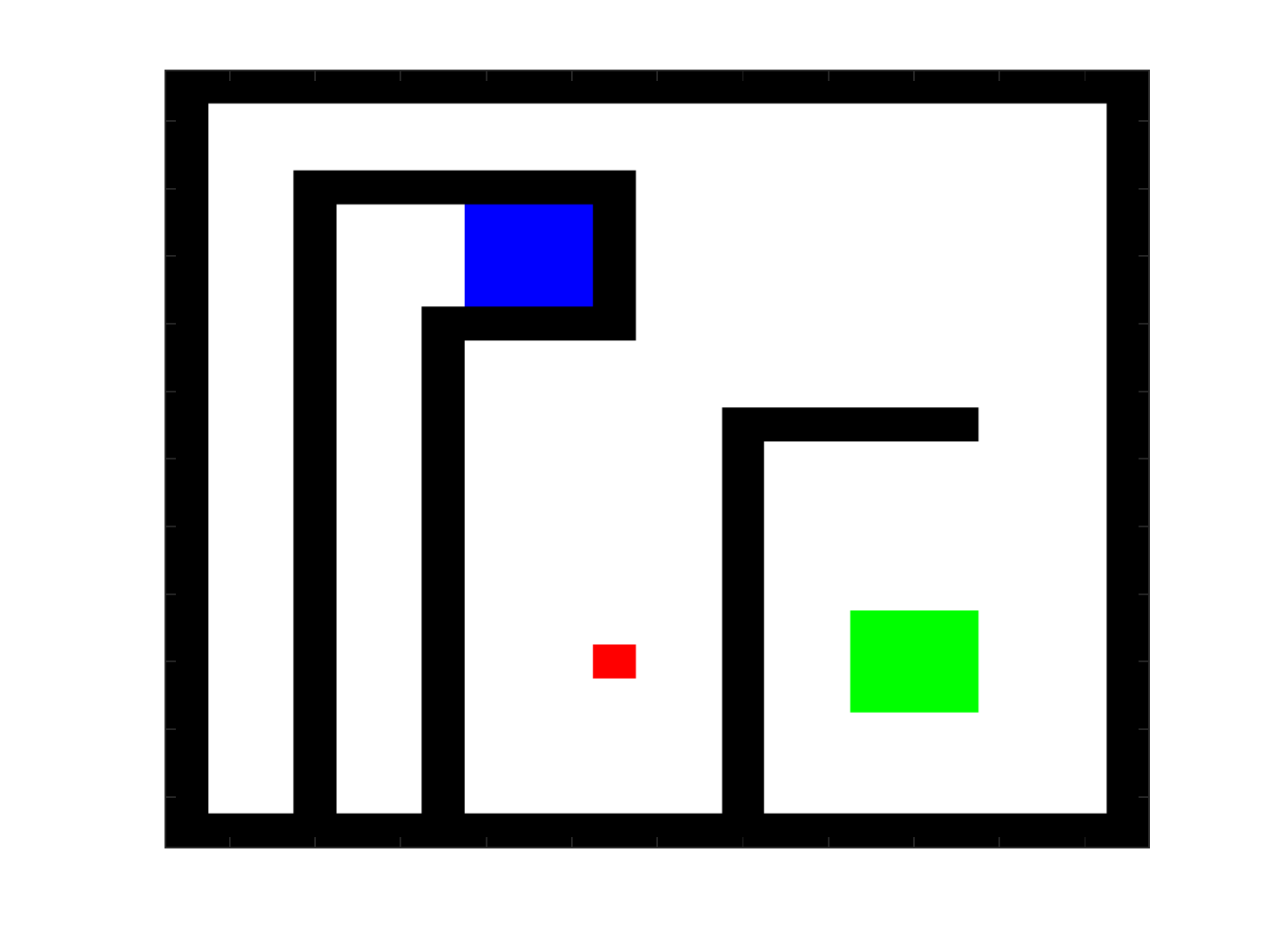} \label{fig:varEnv21}} \subfloat[ENV2 door opened]{\includegraphics[width=0.5\columnwidth]{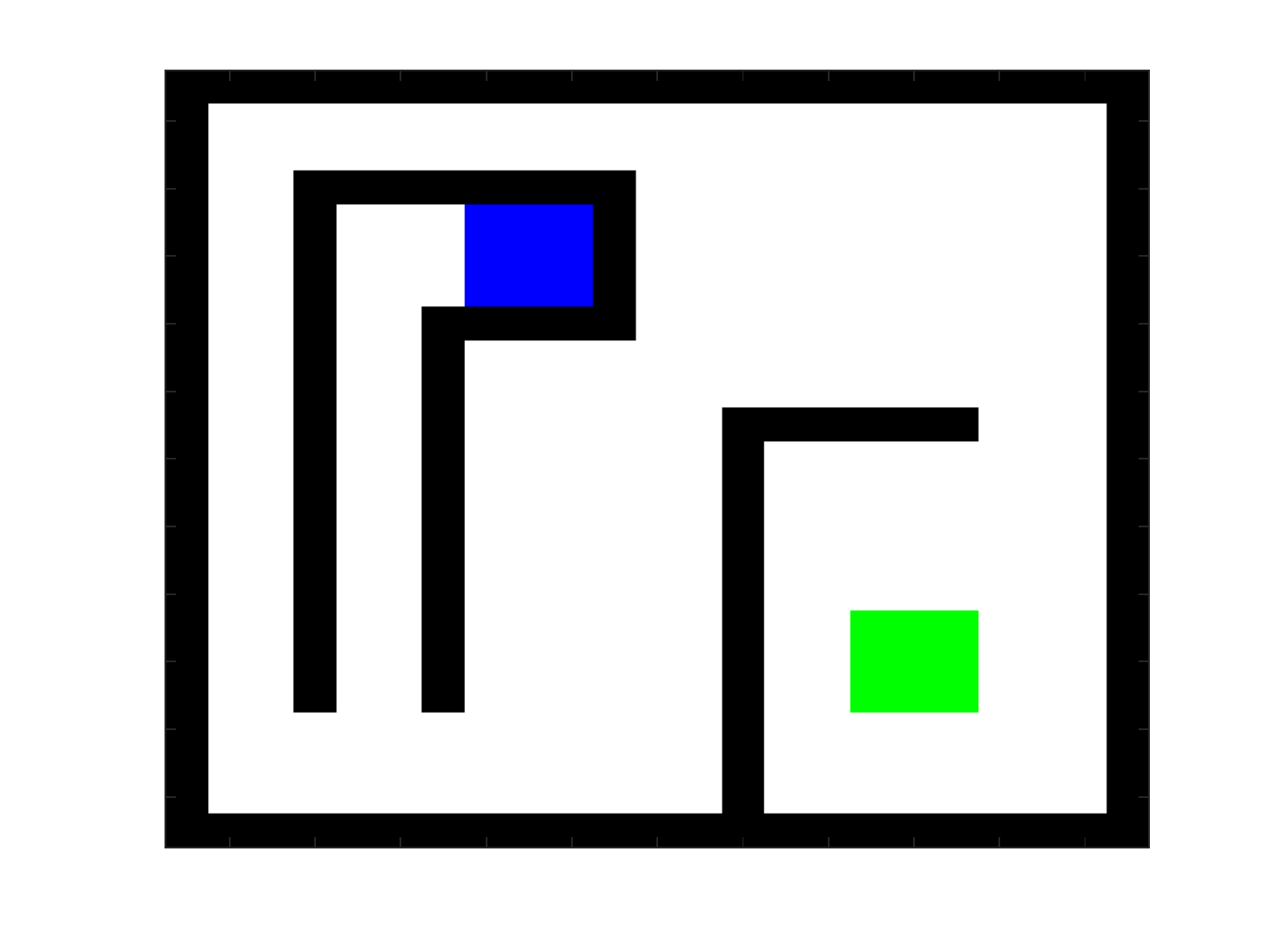} \label{fig:varEnv22}}

\subfloat[ENV3 door closed]{\includegraphics[width=0.5\columnwidth]{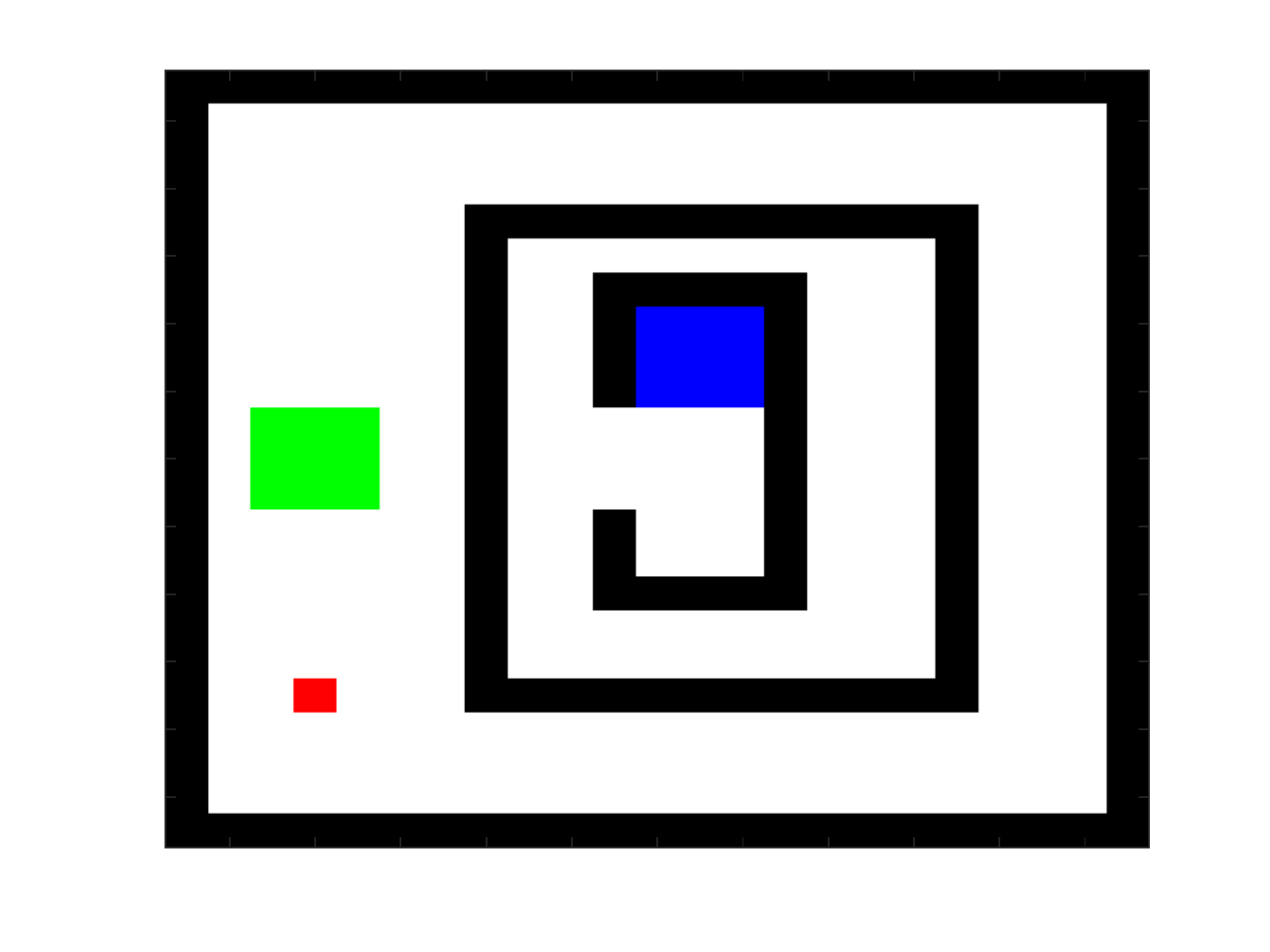} \label{fig:varEnv31}} \subfloat[ENV3 door opened]{\includegraphics[width=0.5\columnwidth]{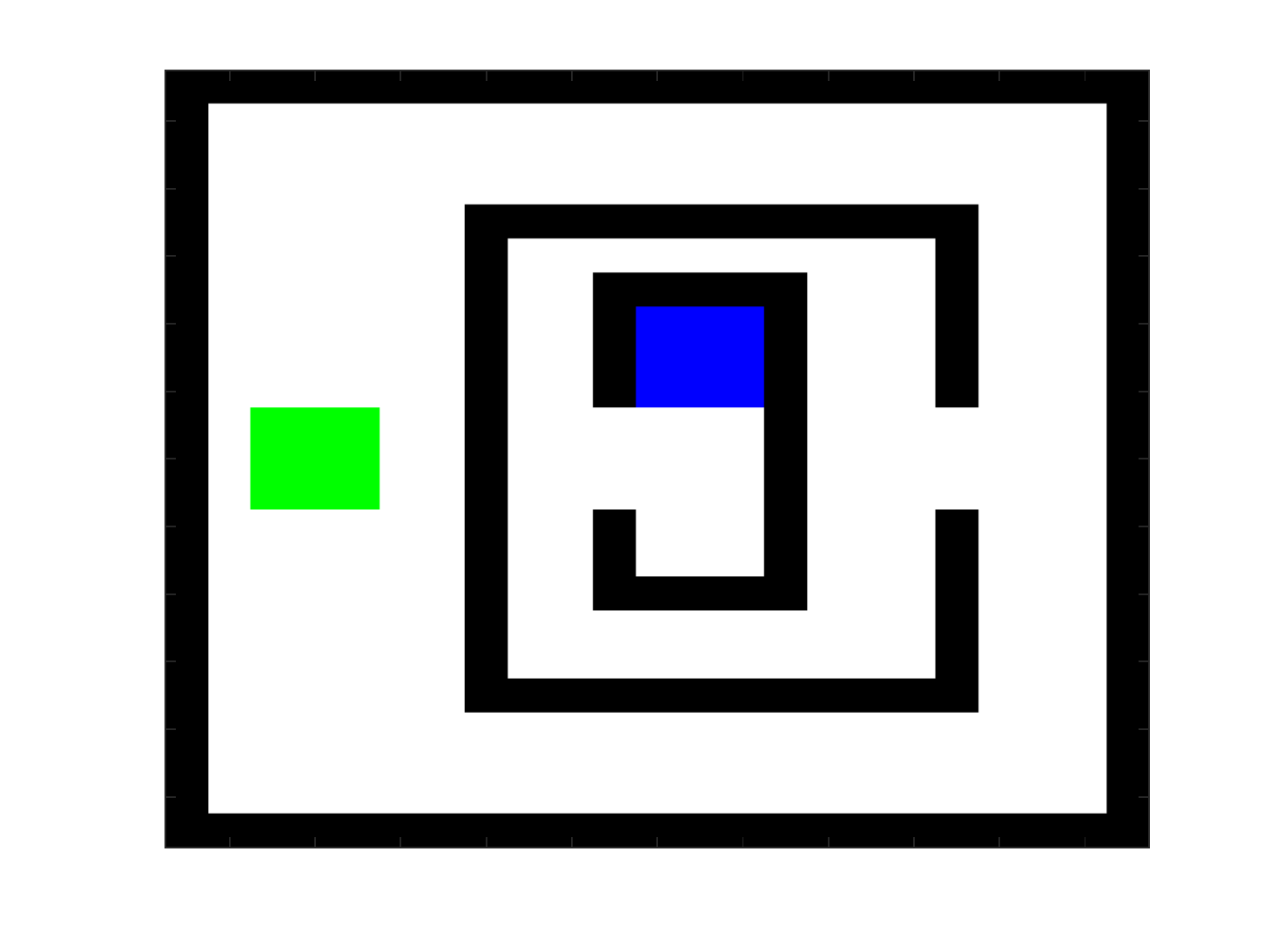} \label{fig:varEnv32}}

\end{subfigures}
\end{center}

\caption{Three additional test environments. Green, blue and red show the button, goal and starting position of the agent respectively. Figures~\ref{fig:varEnv11},~\ref{fig:varEnv21} and \ref{fig:varEnv31} show the versions of the environments where door is closed, while Figures~\ref{fig:varEnv12},~\ref{fig:varEnv22} and \ref{fig:varEnv32} show the versions of the environments where the door leading to the goal is opened.} \label{fig:otherTestEnv}
\end{figure}

Figure~\ref{fig:otherTestEnv} shows three additional environments (referred to as ENV1, ENV2 and ENV3) that we used to perform additional test on the evolved NPSP rules. These environments were not used during the evolutionary process of the NPSP rules. The environments shown in the first column are the versions with the door closed, while those shown in the second column are the versions that the door opened. Green, blue and red areas indicate the button, the goal and the starting position of the agent.

Table~\ref{tab:additionalCompResults} shows the additional experimental results we obtained on the environments shown in Figure~\ref{fig:otherTestEnv}. The rows (corresponding to each environment) labelled as ``Novelty'', ``Distance'', ``Second'' and ``Goal'' show respectively the median percentage of the novel behavior, the shortest distance to the goal, the number of times the agent could access the room where the goal is located, and the number of times reached the goal. Each algorithm was tested on each environment for 25 trials.  

The results are similar to those obtained in the previous experiments. Overall, larger network sizes produced more novel behaviors. Similarly, NPSP0H shows competitive performance even against the network with largest number of hidden neurons (i.e., NPSP50H).

\begin{table*}[]
\centering
\small
\caption{The median of the novelty and distance measures of agents tested on additional environments shown in Figure~\ref{fig:otherTestEnv}.}\label{tab:additionalCompResults}

\begin{tabular}{|l|l|l|l|l|l|l|l|l|l|}  
\hline
{\textbf{Environment}} & {\textbf{Algorithm}}         &
\textbf{RS0H} &\textbf{NPSP0H}  & \textbf{RS15H} &\textbf{NPSP15H}&\textbf{RS30H}&\textbf{NPSP30H} &\textbf{RS50H}&\textbf{NPSP50H} \\ \hline
\multirow{4}{*}{\textbf{ENV1}} &\cellcolor[HTML]{EFEFEF}\textbf{Novelty}   &\cellcolor[HTML]{EFEFEF}0.08& \cellcolor[HTML]{EFEFEF}0.37 &\cellcolor[HTML]{EFEFEF}0.27&\cellcolor[HTML]{EFEFEF}0.34&\cellcolor[HTML]{EFEFEF}0.45&\cellcolor[HTML]{EFEFEF}0.68&\cellcolor[HTML]{EFEFEF}0.63&\cellcolor[HTML]{EFEFEF}1.00\\ \cline{2-10} 
                               & \textbf{Distance}  &1.38 &1.3 &1.38&1.38&1.38&0.78&1.38&0.86\\ \cline{2-10} &\cellcolor[HTML]{EFEFEF}\textbf{Second} &\cellcolor[HTML]{EFEFEF}1&\cellcolor[HTML]{EFEFEF}11&\cellcolor[HTML]{EFEFEF}0&\cellcolor[HTML]{EFEFEF}9&\cellcolor[HTML]{EFEFEF}1&\cellcolor[HTML]{EFEFEF}18&\cellcolor[HTML]{EFEFEF}8&\cellcolor[HTML]{EFEFEF}17\\ \cline{2-10} 
                               & \textbf{Goal}   & 0 &7 &0&2&0&1&0&4\\ \hline\hline

\multirow{4}{*}{\textbf{ENV2}} & \cellcolor[HTML]{EFEFEF}\textbf{Novelty}   &\cellcolor[HTML]{EFEFEF}0.10&\cellcolor[HTML]{EFEFEF}0.43 &\cellcolor[HTML]{EFEFEF}0.29&\cellcolor[HTML]{EFEFEF}0.48&\cellcolor[HTML]{EFEFEF}0.47&\cellcolor[HTML]{EFEFEF}0.74&\cellcolor[HTML]{EFEFEF}0.64&\cellcolor[HTML]{EFEFEF}1.00\\ \cline{2-10} 
                               & \textbf{Distance}  & 1.30& 0.94&1.30&1.30&1.30&1.30&1.30&1.30\\ \cline{2-10}  &\cellcolor[HTML]{EFEFEF}\textbf{Second} &\cellcolor[HTML]{EFEFEF}1&\cellcolor[HTML]{EFEFEF}14&\cellcolor[HTML]{EFEFEF}0&\cellcolor[HTML]{EFEFEF}0&\cellcolor[HTML]{EFEFEF}0&\cellcolor[HTML]{EFEFEF}1&\cellcolor[HTML]{EFEFEF}0&\cellcolor[HTML]{EFEFEF}0\\ \cline{2-10} 
                               & \textbf{Goal}   & 0&2  & 0 &0&0&0&0&0\\ \hline\hline

\multirow{4}{*}{\textbf{ENV3}} & \cellcolor[HTML]{EFEFEF}\textbf{Novelty}   &\cellcolor[HTML]{EFEFEF}0.09&\cellcolor[HTML]{EFEFEF}0.37&\cellcolor[HTML]{EFEFEF}0.32&\cellcolor[HTML]{EFEFEF}0.35&\cellcolor[HTML]{EFEFEF}0.55&\cellcolor[HTML]{EFEFEF}0.98&\cellcolor[HTML]{EFEFEF}0.73&\cellcolor[HTML]{EFEFEF}1.00\\ \cline{2-10} 
                               & \textbf{Distance}  &1.40& 0 &1.40&1.40&1.40&0.60&0.65&0.60\\ \cline{2-10} 
                               & \cellcolor[HTML]{EFEFEF}\textbf{Second} &\cellcolor[HTML]{EFEFEF}1&\cellcolor[HTML]{EFEFEF}20&\cellcolor[HTML]{EFEFEF}3&\cellcolor[HTML]{EFEFEF}11&\cellcolor[HTML]{EFEFEF}6&\cellcolor[HTML]{EFEFEF}19&\cellcolor[HTML]{EFEFEF}18&\cellcolor[HTML]{EFEFEF}20\\ \cline{2-10} 
                               & \textbf{Goal}   & 1&18&0&2&0&1&0&2\\ \hline
\end{tabular}
\end{table*}


\section{Conclusions and Future Work}\label{sec:Conclusions}

In this work, we proposed using synaptic plasticity to allow learning in ANNs for the cases where there is no fitness value or reinforcement signals. We refer to those problems as the ``needle in a haystack'' due to the difficulty of finding the solutions in a large search space. We proposed an evolutionary approach dubbed as novelty producing synaptic plasticity (NPSP), whose goal is to produce as many novel behaviors as possible and find the behavior that can solve the problem. The NPSP performs synaptic changes based on a data structure (neuron activation traces) that stores pairwise activations of neurons during an episode. We compared the NPSP with random search and random walk algorithms that are analogous to the NPSP except that they perform synaptic changes randomly. Our results show that the information about the pairwise activations of neurons introduced with the NATs helps increase the number of novel behaviors relative to random search and random perturbations.

We tested our algorithms on complex maze-navigation tasks where defining the fitness function is not straightforward. We observed a positive relation between producing novel behaviors and finding a solution in these tasks. We also investigated the generalizability of the NPSP rule by testing them for different starting points and in different environments that were not used for the training. In some starting points/environments, the NPSP was not able to produce as many novel behaviors as it produced in others.

We performed experiments on recurrent neural networks with various sizes. We observed that the networks with a larger number of hidden neurons produced more novel behaviors. However, this did not directly cause a higher chance of finding the goal position. This may be due to the capability of large networks to produce more complex behaviors which may not necessarily lead to efficient (i.e. goal-reaching) exploration behavior patterns in the environment.

There are several interesting research questions we aim to follow starting from this work. First, we may consider the fact that we are not necessarily interested in finding all behavioral patterns, because many of these behaviors may not make sense. For instance, if we want to explore the environment, going front and back or cycling around would not help. It would be interesting to introduce some sort of bias, or constraints, in the generation of certain types of behaviors. However, this may also restrict the search and prevent finding good solutions so a way to guarantee a good compromise between solution novelty and solution efficiency should be investigated.

Second, it may be interesting to use multi-objective optimization to select the NPSP rules based also on their capability of getting closer to the goal. However, this may introduce an environmental bias (this is the main reason we did not use it already this work). To avoid that, the rules may be required to be evaluated in many different environments.

Another interesting research question concerns the synaptic adjustments. Especially in large networks, small adjustments in connections may add up to large behavioral changes. It would be interesting to investigate how to perform these changes to allow behavioral continuity.

Finally, evolutionary computation is a powerful tool to discover different plasticity mechanisms in various learning scenarios. It may be interesting to investigate different plasticity mechanisms and see how they perform synaptic adjustments.



\end{document}